\documentclass[10pt,twocolumn,letterpaper]{article}

\usepackage{cvpr}      
\usepackage{pifont}
\usepackage{multirow}
\definecolor{cvprblue}{rgb}{0.21,0.49,0.74}
\usepackage[pagebackref,breaklinks,colorlinks,allcolors=cvprblue]{hyperref}
\usepackage{xcolor}
\usepackage[table]{xcolor}
\def\paperID{540} 
\def\confName{CVPR}
\def\confYear{2025}

\definecolor{mygray}{gray}{0.87}
\definecolor{first}{RGB}{255,179,179}
\definecolor{second}{RGB}{255,255,179}
\newcommand\ChangeRT[1]{\noalign{\hrule height #1}}
\newcommand{\cmark}{\ding{51}}
\newcommand{\xmark}{\ding{55}}
\title{\textbf{{\color[RGB]{240,161,172}R}{\color[RGB]{245, 194, 66}E}{\color[RGB]{240, 204, 177}A}{\color[RGB]{155, 229, 201}L}{\color[RGB]{100, 192, 229}M}}: An MLLM-Agent Framework for Open World 3D Reasoning Segmentation and Editing on Gaussian Splatting}

\author{
    Changyue Shi$^{1,2*}$
    \quad Minghao Chen$^{1*}$
    \quad Yiping Mao$^{1}$
    \quad Chuxiao Yang$^{1}$ \\
    Xinyuan Hu$^{1}$
    \quad Jiajun Ding$^{1\dagger}$
    \quad Zhou Yu$^{1}$
    \vspace{1em}
    \\
    $^{1}$Hangzhou Dianzi University
    \quad
    $^{2}$Peking University
}

\begin{document}

\twocolumn[{%
\renewcommand\twocolumn[1][]{#1}
\maketitle
\vspace{-5pt}
\vspace{-3em}
\begin{center}
    \captionsetup{type=figure}
    \includegraphics[width=0.9\linewidth]{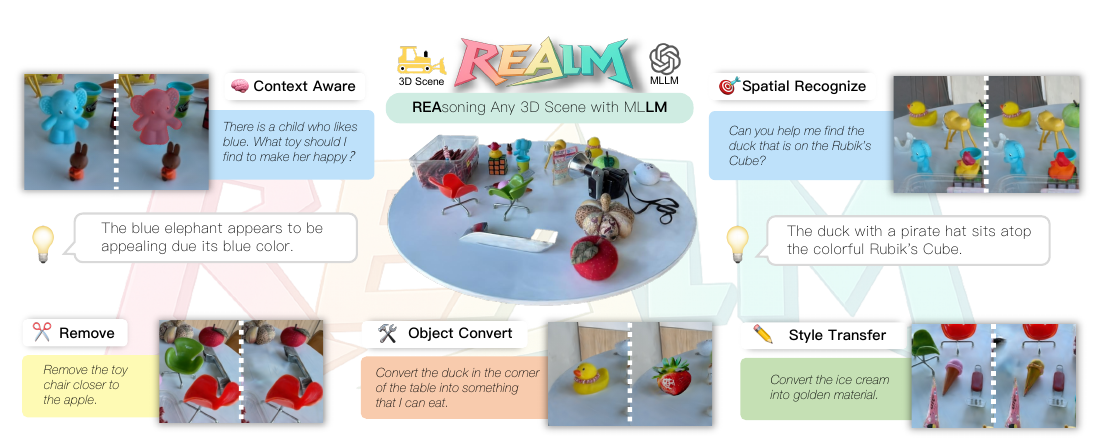}
    \vspace{-7pt}
    \caption{We propose \textbf{{\color[RGB]{240,161,172}R}{\color[RGB]{245, 194, 66}E}{\color[RGB]{240, 204, 177}A}{\color[RGB]{155, 229, 201}L}{\color[RGB]{100, 192, 229}M}}, an MLLM-agent framework designed for open-world 3D reasoning segmentation and editing within 3D Gaussian Splatting (3DGS). REALM can perform reasoning over implicit instructions and accurately segment the target object. REALM also supports various 3D editing instructions, including object removal, replacement, and style transfer.
    }
    \label{fig:teaser}
\end{center}
%
}]

\def\thefootnote{*}\footnotetext[1]{\footnotesize Equal contribution: Changyue Shi and Minghao Chen.}
\def\thefootnote{$\dagger$}\footnotetext[2]{\footnotesize Corresponding author: Jiajun Ding (\href{mailto:djj@hdu.edu.cn}{djj@hdu.edu.cn}).}

\begin{abstract}
Bridging the gap between complex human instructions and precise 3D object grounding remains a significant challenge in vision and robotics. Existing 3D segmentation methods often struggle to interpret ambiguous, reasoning-based instructions, while 2D vision-language models that excel at such reasoning lack intrinsic 3D spatial understanding. In this paper, we introduce \textbf{{\color[RGB]{240,161,172}R}{\color[RGB]{245, 194, 66}E}{\color[RGB]{240, 204, 177}A}{\color[RGB]{155, 229, 201}L}{\color[RGB]{100, 192, 229}M}}, an innovative MLLM-agent framework that enables open-world reasoning-based segmentation without requiring extensive 3D-specific post-training. We perform segmentation directly on 3D Gaussian Splatting representations, capitalizing on their ability to render photorealistic novel views that are highly suitable for MLLM comprehension. As directly feeding one or more rendered views to the MLLM can lead to high sensitivity to viewpoint selection, we propose a novel Global-to-Local Spatial Grounding strategy. Specifically, multiple global views are first fed into the MLLM agent in parallel for coarse-level localization, aggregating responses to robustly identify the target object. Then, several close-up novel views of the object are synthesized to perform fine-grained local segmentation, yielding accurate and consistent 3D masks. Extensive experiments show that REALM achieves remarkable performance in interpreting both explicit and implicit instructions across LERF, 3D-OVS, and our newly introduced REALM3D benchmarks. Furthermore, our agent framework seamlessly supports a range of 3D interaction tasks, including object removal, replacement, and style transfer, demonstrating its practical utility and versatility. 
Project page: \url{https://ChangyueShi.github.io/REALM}.
\end{abstract}    
\section{Introduction}
\vspace{5pt}
\begin{minipage}{\linewidth}
\noindent
\emph{``Vision is the process of discovering from images what is present in the world, and where it is.''} 

\hfill --- David Marr (1982)
\end{minipage}
\vspace{5pt}

Endowing AI agents with the ability to understand and interact with the 3D world through natural language is a cornerstone for the future of robotics and human-AI collaboration. Humans effortlessly perform complex instructions by first interpreting the request and then grounding it in their spatial surroundings. For instance, when given the instruction \textit{``make the table tidier''}, a person will first identify both the storage container and the loose objects, then gather and place the clutter appropriately. The crucial first step is to accurately segment target objects based on implicit, commonsense reasoning. While this is naturally for humans, achieving such reasoning-based 3D segmentation remains a challenge for current AI agents \cite{kerr2024robot, zheng2024gaussiangrasper}.

\begin{figure}[t]
    \centering
    \includegraphics[width=0.87\linewidth]{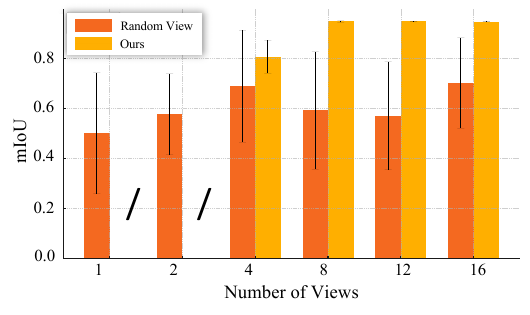}
    \vspace{-10pt}
    \caption{\textbf{REALM vs. Direct Image Inputs.} Feeding one or a few random rendered views into the MLLM makes the outcome highly sensitive to viewpoint selection.
    } 
    \vspace{-15pt}
    \label{fig:random_vs_ours}
\end{figure}

Existing research streams offer partial but incomplete solutions. On the one hand, 3D open-vocabulary segmentation methods have made strides in linking language to 3D representations, such as point cloud~\cite{huang2025reason3d}, NeRFs \cite{kerr2023lerf} or 3D Gaussian Splatting (3DGS) \cite{qin2024langsplat}. However, they primarily excel at explicit, direct queries (e.g., ``segment the cup'') and falter when faced with instructions that demand reasoning about spatial relationships, semantic attributes, or common knowledge (e.g., ``segment the object between the lamp and the book''). 
On the other hand, Multimodal Large Language Models (MLLMs) \cite{bai2025qwen2, li2023blip, liu2023visual} have demonstrated success in 2D visual reasoning~\cite{imp2024, lai2024lisa, fu2023guiding}. Pretrained on large-scale 2D image–text datasets, MLLMs 
can interpret ambiguous instructions with remarkable accuracy, but typically lack 3D spatial awareness and the ability to precisely ground their findings in space. Earlier attempts, including ScanReason~\cite{zhu2024scanreason} and VGMamba~\cite{zhu2025vgmamba}, are limited to predicting 3D bounding boxes rather than the fine-grained masks we pursue. While ReasonGrounder~\cite{liu2025reasongrounder} is conceptually closer to our task, it relies heavily on a top-down view, which limits its applicability in complex 3D environments.


In this paper, we propose \textbf{REALM} to bridge this gap by leveraging the powerful reasoning capabilities of off-the-shelf MLLMs for 3D segmentation. We adopt 3DGS \cite{kerbl20233d} as a high-fidelity proxy for the 3D world, capitalizing on its ability to render photorealistic novel views that are perfectly suited for MLLM comprehension. In the REALM framework, we first optimize a \textit{3D Feature Field} that can assign an identity feature to each Gaussian primitive. Next, we introduce \textit{MLLM‐based Instance Segmenter (LMSeg)} to perform image-level reasoning segmentation. \textit{LMSeg} generates semantic masks by combining priors from an MLLM~\cite{bai2025qwen2} and SAM~\cite{kirillov2023segment}. These 2D masks are then linked back to their corresponding Gaussian identities in the feature field.

However, feeding a single rendered view to the MLLM is highly sensitive to viewpoint selection: A suboptimal view may obscure the target object or fail to provide sufficient context. Conversely, inputting numerous views simultaneously overwhelms the MLLM, which struggles to resolve ambiguities and establish a consistent 3D understanding (demonstrated in Fig.~\ref{fig:random_vs_ours}). To aggregate multiview results, we propose \textit{Global‐to‐Local Spatial Grounding (GLSpaG)}. 
In the global stage, MLLM agents survey the scene from multiple, diverse viewpoints in parallel, aggregating responses to form a coarse-level localization of the target object.
In the local stage, the agents synthesize several close-up views centered on the identified object and perform fine-grained segmentation.
Once the instance is segmented in 3D space, REALM can execute a range of 3D interaction tasks, e.g., object removal, object replacement, and style transfer, as shown in Fig.~\ref{fig:teaser}.

Since existing benchmarks for 3D segmentation primarily feature explicit prompts, they are inadequate for evaluating performance on reasoning-based tasks. To address this, we re-annotate prominent datasets like LERF \cite{kerr2023lerf} and 3D-OVS~\cite{liu2023weakly} with implicit, reasoning-based instructions. Furthermore, to catalyze future research, we introduce REALM3D, a new large-scale benchmark comprising hundreds of complex scenes along with reconstructed 3DGS and thousands of high-quality, both reasoning-based and non-reasoning-based prompt-mask pairs.



Our contributions can be summarized as follows:
\begin{itemize}
    \item We propose REALM, an MLLM-agent framework for 3D reasoning segmentation, which leverages 3DGS as a proxy to lift the 2D reasoning capability of MLLMs into the 3D domain. Furthermore, REALM supports downstream object-level interactions within 3D scenes through complex textual instructions.

    \item In REALM, \textit{MLLM-Based Instance Segmenter} is proposed to perform image-level reasoning segmentation and infer the corresponding Gaussian identity. To produce accurate 3D object masks, we propose \textit{Global-to-Local Spatial Grouding}, which aggregates image-level reasoning segmentations in a global-to-local manner.
    
    \item We re-annotate LERF and 3D-OVS datasets with implicit queries. We further introduce the REALM3D dataset for evaluating 3D reasoning segmentation, comprising 100+ scenes and 1000+ implicit prompt–mask pairs.

\end{itemize}
\begin{figure*}[t]
    \centering
    \includegraphics[width=0.95\linewidth]{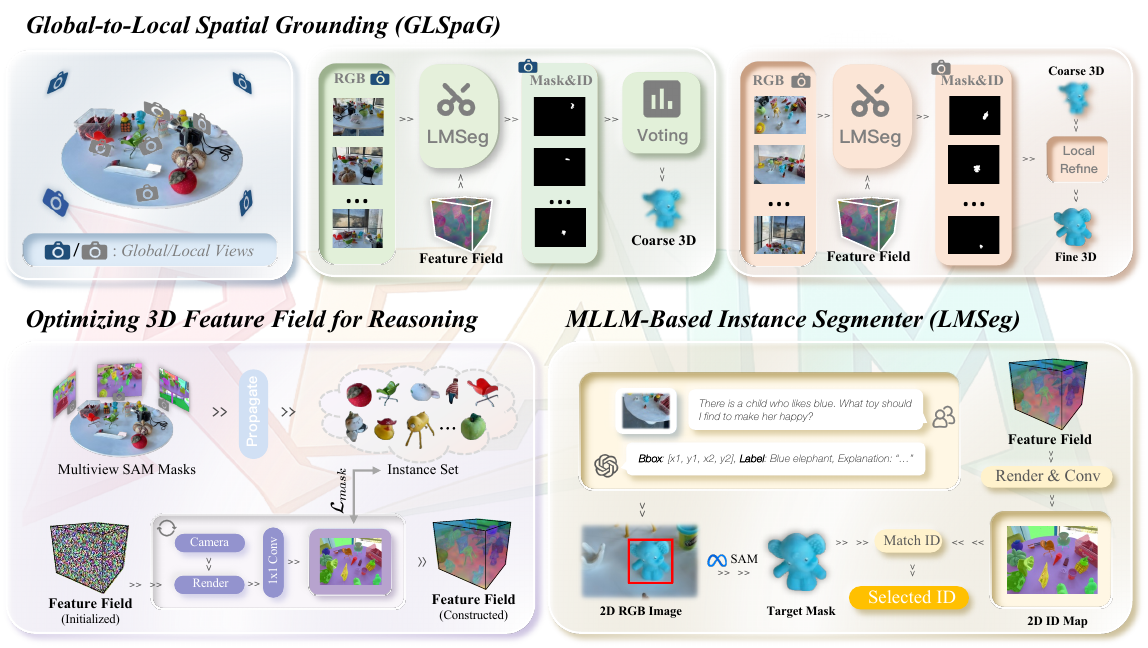}
    \caption{\textbf{Overview of \textbf{{\color[RGB]{240,161,172}R}{\color[RGB]{245, 194, 66}E}{\color[RGB]{240, 204, 177}A}{\color[RGB]{155, 229, 201}L}{\color[RGB]{100, 192, 229}M}}.} 
    \textit{Top:} \textit{Global-to-Local Spatial Grounding (GLSpaG)} pipline hierarchically aggregates the outputs of \textit{LMSeg} agents from global context to local refinement.
    \textit{Bottom left:} We optimize a 3D feature field from 2D SAM masks for 3D consistent identification.   \textit{Bottom right:} \textit{MLLM-based Visual Segmenter (LMSeg)} performs image-level reasoning on one viewpoint and integrates identity information from the optimized feature field to determine the selected instance ID.
    }
    \vspace{-5pt}
    \label{fig:pipeline}
\end{figure*}

\section{Related Works}

\subsection{3D Scene Representations} 
A fundamental step in understanding a 3D scene is to first establish a 3D scene representation. Traditional methods such as Structure-from-Motion (SfM)~\cite{ullman1979interpretation} and Multi-View Stereo (MVS)~\cite{tomasi1992shape} rely on geometric reconstruction techniques. Neural Radiance Field (NeRF)~\cite{mildenhall2021nerf} introduces a learning-based approach. While subsequent NeRF-based methods~\cite{muller2022instant, chen2022tensorf} have enhanced rendering quality and efficiency of the vanilla NeRF, they remain constrained by the computational overhead of volumetric rendering.
Gaussian Splatting~\cite{kerbl20233d} has emerged as an efficient alternative, leveraging rasterization to achieve real-time, high-fidelity scene reconstruction. The representation of 3D Gaussians has inspired extensive researches across various domains, including few-shot reconstruction~\cite{shi2025mmgs, hu2025texture, shi2025sparse4dgs}, super-resolution reconstruction~\cite{feng2024srgs, hu2025srsplat, yang2025sr4d, feng2025ie}, language embedding~\cite{peng2024gags, qin2024langsplat}, and 3D segmentation~\cite{lyu2024gaga}, among others.

\subsection{3D Open-World Understanding} 
Recent research has explored various strategies to incorporate 2D semantic features into 3D representations for enhanced scene understanding. LERF~\cite{kerr2023lerf} pioneers the idea of embedding CLIP features into radiance fields. Subsequent works~\cite{qin2024langsplat, zhou2024feature} leverage 3D Gaussian Splatting (3DGS)~\cite{kerbl20233d} to improve the efficiency of open-vocabulary 3D scene querying. Other approaches lift 2D masks predicted by SAM~\cite{kirillov2023segment} into 3D space. Garfield~\cite{kim2024garfield} and SAGA~\cite{cen2025segment} employ contrastive learning to enable multi-scale instance segmentation. GS-Grouping~\cite{ye2024gaussian} introduces an unsupervised 3D regularization loss to improve performance. In these methods, grouped 3D instances can be queried via 2D prompts. However, existing methods are not capable of handling implicit natural language instructions.

\subsection{Multimodal Large Language Models}
Inspired by the success of large language models (LLMs)~\cite{brown2020language, raffel2020exploring}, recent research has extended their capabilities to process and reason over multiple modalities, including vision and language~\cite{ding2025gpt4image}.
Early work~\cite{fang2024learnability, fang2022out} such as CLIP~\cite{radford2021learning} focused on learning aligned image-text representations for retrieval and classification. 
Subsequent models like Flamingo~\cite{alayrac2022flamingo} and BLIP-2~\cite{li2023blip} introduced lightweight vision-language bridging modules on top of frozen language models, enabling zero-shot image captioning and visual question answering.
More recently, general-purpose MLLMs such as GPT-4V~\cite{achiam2023gpt} and Qwen-2.5-VL~\cite{bai2025qwen2} have demonstrated strong multimodal reasoning abilities.
These models unify textual and visual information within a single autoregressive framework, enabling coherent reasoning across modalities.
In this work, we further explore the multimodal reasoning capabilities of MLLMs in the context of 3D visual grounding.

\begin{figure}
    \centering
    \includegraphics[width=0.9\linewidth]{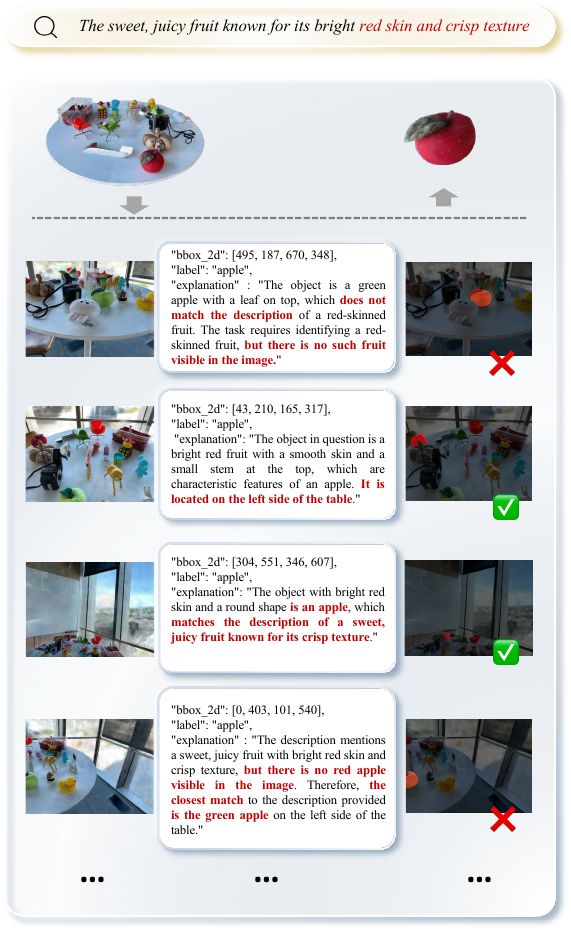}
    \vspace{-5pt}
    \caption{\textbf{Global reasoning process.} We visualize reasoning outputs of the MLLM for each global view. 
    } 
    \vspace{-10pt}
    \label{fig:q_and_a}
\end{figure}

\section{Methodology}
The overview is illustrated in Fig.~\ref{fig:pipeline}. In Sec.~\ref{sec:gs-group}, we construct a 3D instance field for consistent identification. 
In Sec.~\ref{sec:LMSeg}, we introduce an MLLM-agent named \textit{MLLM-Based Visual Segmenter (LMSeg)} to perform image-level reasoning and grounding.
In Sec.~\ref{sec:glspag1} and Sec.~\ref{sec:glspag2}, we introduce the overall agent framework \textit{Global-to-Local Spatial Grounding (GLSpaG)} that aggregates the results of multi-view reasoning segmentation in a gobal-to-local manner.

\subsection{3D Feature Field for Reasoning}
\label{sec:gs-group}
REALM utilizes the proxy of 3DGS~\cite{kerbl20233d} to perform 3D reasoning segmentation. 3DGS~\cite{kerbl20233d} models the scene as a collection of 3D Gaussian primitives. Following previous work~\cite{ye2024gaussian}, we construct a feature field that clusters Gaussian primitives for subsequent 3D reasoning segmentation. 

We first utilize SAM to extract instance masks for each input image. We employ a temporal propagation model~\cite{cheng2023tracking} to associate instances across views. This process ensures that each instance is assigned a consistent identity $id_i$ across all views. To group 3D Gaussians into instances, we assign each Gaussian $G_i = \left \{ x_i, s_i, r_i, o_i, c_i \right \}$ with an instance feature $f_i \in \mathbb{R}^{D}$. The feature can be rendered to a 2D feature map via alpha blending:
\begin{equation}  
\label{eq:render_feature}
F=\sum_{i=1}^{n}f_{i}\alpha _{i}\prod_{j=1}^{i-1}(1-\alpha _{j}).
\end{equation}  

We then apply a classifier \(\mathcal CLS\) to the rendered feature map \(F\) to directly compute the pixel-wise identity map:
\begin{equation}
\label{eq:2dconv}
\hat{{id}}(u,v) = \arg\max_{k}\,\bigl({CLS}(F)_{u,v,k}\bigr),
\end{equation}
where \(\hat{{id}}(u,v)\) denotes the predicted instance ID at pixel \((u,v)\), and \(k\) indexes the instance categories. The Gaussians and the classifier can be supervised by aligning $\hat{{id}}$ with ${{id}}$. After optimization, the trained classifier can be directly applied to the instance features of 3D Gaussians, allowing them to be grouped into their corresponding instances.



\begin{figure*}
    \centering
    \includegraphics[width=0.9\linewidth]{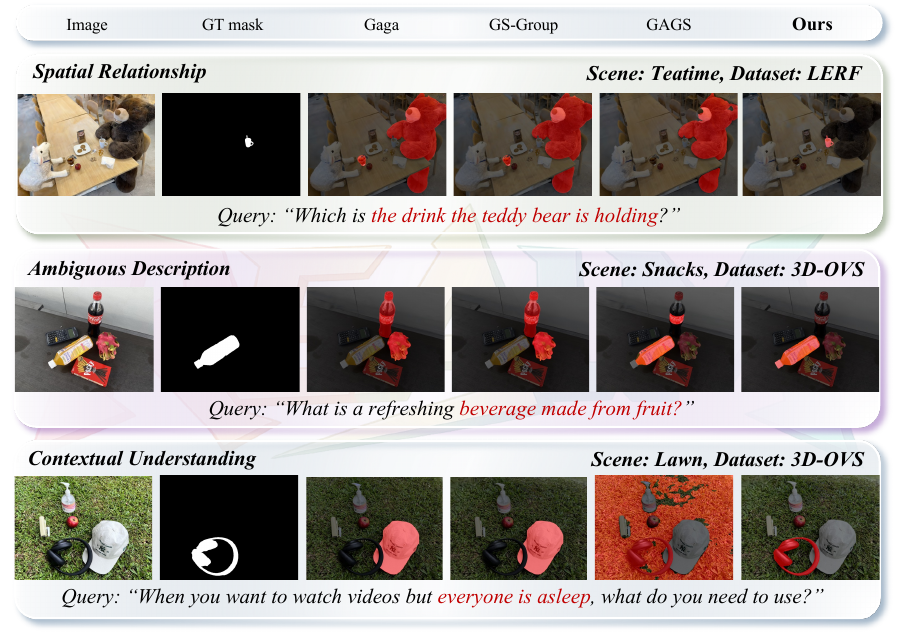}
    \vspace{-5pt}
    \caption{\textbf{Qualitative Results on the LERF Dataset.}
The results demonstrate the ability of REALM to handle complex and implicit language queries with accurate visual grounding.
    } 
    \vspace{-5pt}
    \label{fig:vis_lerf}
\end{figure*}

\subsection{MLLM-Based Visual Segmenter (LMSeg)}
\label{sec:LMSeg}
In this section, we introduce the MLLM-Based Visual Segmenter Agent. With the semantic priors of MLLM, it reasons the implicit queries and outputs the target instance ID using the constructed feature field. Specifically, given an image $\mathcal{I}$ from an arbitrary viewpoint $\mathcal{\phi}$ and a language query $q$, \textit{LMSeg} employs a prompt engineering technique to query an MLLM and returns the following attributes:
\begin{equation}
    (\mathcal{B},\ \mathcal{C},\ \mathcal{E}) = \text{MLLM}(\mathcal{I},\ q),
\end{equation}
where $\mathcal{B} = \{(x_1, y_1, x_2, y_2)\}$ represents the predicted 2D bounding box coordinates, $\mathcal{C}$ denotes the object category, and $\mathcal{E}$ is a concise explanatory rationale. The predicted bounding box $\mathcal{B}$ is subsequently fed into SAM~\cite{kirillov2023segment} to generate the corresponding binary object mask $M^{2D} \in \left \{0,1 \right \}^{H \times W}$, where each element indicates whether the pixel belongs to the target object.

With the constructed feature field $G$ and the trained classifier $CLS$ described in Sec.~\ref{sec:gs-group}, we infer the 2D instance map $\hat{{id}}$ at viewpoint $\phi$ using Eq.~\ref{eq:render_feature} and~\ref{eq:2dconv}. By intersecting the binary mask $M$ with the predicted instance map $\hat{\mathrm{id}}$, we reliably identify the target instance $ID$ at viewpoint $\phi$. 


\subsection{Global-to-Local Spatial Grounding (Global)}
\label{sec:glspag1}
Feeding a single rendered view to the MLLM is highly sensitive to viewpoint selection. To address this, we first sample a set of global viewpoints. For each view, we apply \textit{LMSeg} to infer the target instance identity. These per-view instance IDs are then aggregated and used to group the target 3D Gaussians within the constructed 3D feature field. We visualize the process in Fig.~\ref{fig:q_and_a}

\noindent \textbf{\textit{Global Cameras}} Given the training camera set $\phi^{\text{train}}$, the sampling of global viewpoints should adhere to the following principles: 1) Covering diverse spatial locations., 2) Covering multiple objects with minimal views. To achieve this, we cluster the training camera poses using K-means and select one representative camera from each cluster:
\begin{equation}
    \{\phi^{\text{cluster}}_i\}_{i=1}^{N^{\text{cluster}}} = \text{KMeans}(\{\phi^{\text{train}}_j\}_{j=1}^{N^{\text{train}}},\ N^{\text{cluster}}).
\end{equation}

For each view $\phi_i \in \{\phi^{\text{cluster}}_i\}_{i=1}^{N^{\text{cluster}}}$, we compute the number of unique instance IDs in the predicted 2D instance map $\hat{\mathrm{id}}_i$.
Then, we select the top $N^{\text{global}}$ views with the highest instance counts to obtain the global view set:
\begin{equation}
    \{\phi^{\text{global}}_i\}_{i=1}^{N^{\text{global}}} = \text{TopK-ID}\left( \{\phi^{\text{cluster}}_i, \hat{\mathrm{id}}_i\}_{i=1}^{N^{\text{cluster}}},\ N^{\text{global}} \right),
\end{equation}
where \text{TopK-ID} returns the subset of views with the highest number of distinct instance identities.

\begin{table*}[t]
    \centering 
    \resizebox{0.7\textwidth}{!}{
    \renewcommand\arraystretch{1.3}
    \begin{tabular}{l|cc|cc|cc}
        \multirow{2}{*}{Methods} & \multicolumn{2}{c|}{LERF} & \multicolumn{2}{c}{3D-OVS} & \multicolumn{2}{c}{REALM3D}\\
        \cline{2-7}
         & mIoU (\%) $\uparrow$ & mBIoU (\%) $\uparrow$  & mIoU (\%) $\uparrow$ & mBIoU (\%) $\uparrow$ & mIoU (\%) $\uparrow$ & mBIoU (\%) $\uparrow$  \\
\ChangeRT{1.2pt}     
        Gaga~\cite{lyu2024gaga} & \cellcolor{second}44.82 & \cellcolor{second}42.37  & 42.53& 37.38 &58.56&49.65\\
        GAGS~\cite{peng2024gags}& 17.84 & 15.87  &  \cellcolor{second}58.46& \cellcolor{second}50.34 &52.24&39.76\\
        GS-Group~\cite{ye2024gaussian}  & 42.43 & 40.01  & 41.79& 38.28&\cellcolor{second}65.55&\cellcolor{second} 55.99\\
        \textbf{REALM (Ours)}           & \cellcolor{first}\textbf{92.88} &\cellcolor{first} \cellcolor{first}\textbf{90.12}  &\cellcolor{first} \textbf{93.68}& \cellcolor{first}\textbf{86.02} &\cellcolor{first} \textbf{82.30}&\cellcolor{first} \textbf{70.37}\\
    \end{tabular}
    }
        \caption{\textbf{Quantitative results on LERF~\cite{kerr2023lerf}, 3D-OVS~\cite{liu2023weakly} and our proposed REALM3D benchmarks.} We compare REALM with other models on implicit queries. 
        The best results are marked in \textbf{bold}.}
    \label{tab:comp}
\end{table*}

\noindent \textbf{\textit{Global Spatial Grounding}} Once the global cameras are determined, we apply \textit{LMSeg} (see Sec.~\ref{sec:LMSeg}) to each selected view $\phi^{\text{global}}_i$ under the query $q$ to obtain the corresponding 2D instance identity map $ID_i^q$. These ID predictions are then aggregated through a voting scheme to determine the final target instance identity $ID^q \;=\;\underset{c\in\mathcal{C}}{\arg\max}\;\bigl|\{\,i : ID_i^q = c\}\bigr|$, where \(\mathcal{C}\) is the set of all candidate instance IDs.

We utilize the classifier $CLS$ to predict the semantic identity of each Gaussian in the 3D space based on feature $f_i$, thereby producing a 3D segmentation mask $M^{3D}$:
\begin{equation}
M^{3D}_{i} = \begin{cases}1,  \arg\max_{k}\,\bigl({CLS}(f_i)\bigr)=ID^y
 \\0, \arg\max_{k}\,\bigl({CLS}(f_i)\bigr)\ne ID^y
\end{cases}.
\end{equation}

This process yields a coarse 3D segmentation mask, which will be further refined in the subsequent stage.

\subsection{Global-to-Local Spatial Grounding (Local)}
\label{sec:glspag2}
Local grounding samples a set of local cameras and uses fine-grained multi-view 2D masks to refine the coarse 3D mask produced in the global stage. 

\noindent\textbf{\textit{Local Cameras}} Local cameras are sampled from clustered representative cameras $\{\phi^{\text{cluster}}_i\}_{i=1}^{N^{\text{cluster}}}$. A view is selected if the target $ID^y$ appears in its 2D instance map $\hat{\mathrm{id}}_i$:
\begin{equation}
\left\{\phi^{\text{local}}_i\right\}_{i=1}^{N^{\text{local}}} = \left\{ \phi^{\text{cluster}}_j \;\middle|\; ID^y \in \hat{\mathrm{id}}_j,\ j=1,\dots,N^{\text{cluster}} \right\}.
\end{equation}

\noindent\textbf{\textit{Local Spatial Grounding}} We first employ \textit{LMSeg} for each image rendered from $\left\{\phi^{\text{local}}_i\right\}_{i=1}^{N^{\text{local}}}$ to obtain a set of local 2D masks $\left\{M^{2D-Local}_i\right\}_{i=1}^{N^{\text{local}}}$. 

Given a local camera $\phi^{\text{local}}_i$, the 3D mask $M^{3D}$ can be rendered to the image plane via differentiable rasterizer. The rendered mask $\hat{M}_i$ can be aligned with the corresponding 2D mask $M^{2D-Local}_i$ extracted from \textit{LMSeg}:
\begin{equation}
\mathcal{L}_{\text{local}} = \left\| \hat{M}_i - M^{2D\text{-Local}}_i \right\|_1.
\end{equation}

This process enables REALM to produce more semantically accurate 3D masks (see Fig.~\ref{fig:glgroup_ab}).

\begin{table}
    \centering
\resizebox{0.46\textwidth}{!}{%
\renewcommand\arraystretch{1.2}
\begin{tabular}{l|ccc}
 & Num. of Scenes & Prompt-Mask Pairs & Implicit Prompts\\
\ChangeRT{1.2pt}
    LERF~\cite{kerr2023lerf} &  5&  36&  \xmark \\
    3DOVS~\cite{liu2023weakly} & 10&  150 & \xmark \\
  \textbf{REALM3D} & \textbf{100} & \textbf{1444} & \textbf{\cmark }

\end{tabular}%
}
    \caption{\textbf{Statistics of 3D segmentation benchmark.} Compared to existing benchmarks, our REALM3D contains more scenes and prompt-mask pairs. Additionally, REALM3D provides implicit prompts for each mask annotation.}
    \vspace{-10pt}

\label{tab:realm3d_dataset_attribute}
\end{table}

\begin{figure*}
    \centering
    \includegraphics[width=0.88\linewidth]{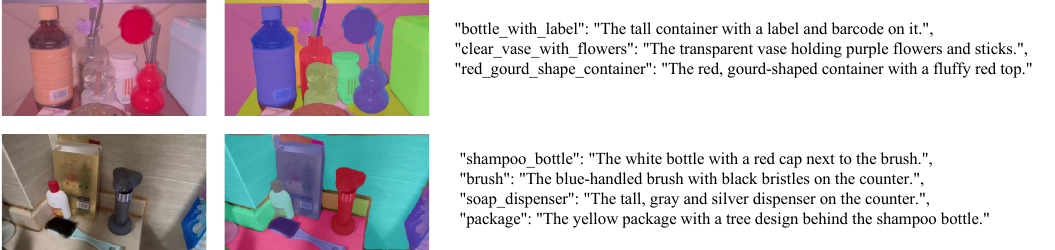}
    
    \caption{\textbf{Examples in REALM3D benchmark.}
We use MLLM~\cite{bai2025qwen2} and SAM~\cite{kirillov2023segment} to annotate over 1K prompt–mask pairs, enabling quantitative evaluation on implicit queries.
    } 
    \label{fig:ream_example}
\end{figure*}

\begin{figure*}[t]
    \centering
    \includegraphics[width=0.88\linewidth]{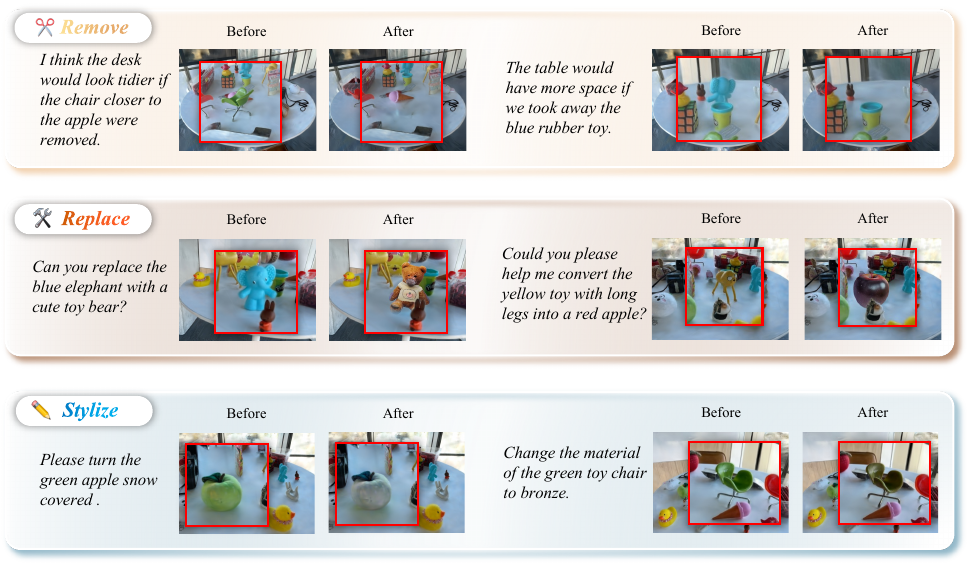}
    \caption{\textbf{Language-driven 3D editing.} Once the object is grounded, we can perform a wide range of 3D editing tasks.
    } 
    \vspace{-5pt}
    \label{fig:edit_results}
\end{figure*}

\section{Experiments}

\subsection{Experimental Settings}
\textbf{Benchmark.} We evaluate REALM and other baselines on LERF~\cite{kerr2023lerf}, 3D-OVS~\cite{liu2023weakly}, and our REALM3D datasets. These datasets cover diverse object layouts and implicit and explicit prompt-mask pairs. 

\noindent  \textit{(1) LERF and 3D-OVS datasets:} We select 2 representative scenes from the LERF dataset and 5 from the 3D-OVS dataset. To establish implicit prompt–mask pairs, we re-annotate the original prompts~\cite{ye2024gaussian} using Qwen2.5-VL and then rigorously manually curate the annotations. 

\noindent \textit{(2) REALM3D dataset:}
To facilitate future research, we introduce REALM3D, a dataset specifically designed to evaluate 3D reasoning segmentation. REALM3D comprises 100+ 3D scenes captured in multiview images, along with 3D point clouds and camera poses generated by VGGT~\cite{wang2025vggt}. We annotate 1k+ prompt–mask pairs using Qwen2.5-VL and SAM, covering diverse forms of implicit and explicit prompts (see Fig.~\ref{fig:ream_example} and Tab.~\ref{tab:realm3d_dataset_attribute}). REALM3D can be used to evaluate the robustness of models across diverse applications. We provide details of REALM3D in the supplementary materials.

\noindent\textbf{Baselines and metrics.} 
We compare REALM against previous state-of-the-art methods for open-vocabulary 3D segmentation, including Gaga~\cite{lyu2024gaga}, GAGS~\cite{peng2024gags}, and GS-Group~\cite{ye2024gaussian}. We report the mIoU and mBIoU following previous works ~\cite{shi2024language, qin2024langsplat, kerr2023lerf, peng2024gags} to quantitatively exanimate the accuracy of 3D reasoning segmentation results.

\noindent\textbf{Implementation.} We implement REALM using the PyTorch framework. We set the number of clustered views $N^{\text{cluster}} = 24$ and the number of global views $N^{\text{global}} = 8$. The local refinement is performed with 50 optimization steps. The selection of these hyper-parameters is further discussed in the ablation study. More implementation details can be found in the supplementary materials. All the results can be obtained using an NVIDIA RTX 3090 GPU.

\begin{table*}[t]
	\renewcommand\arraystretch{1.2}
	\scriptsize
	\begin{subtable}[t]{0.34\textwidth}
		\centering
            \begin{tabular}{ccc}
                Methods & mIoU & mBIoU \\ 
		      \ChangeRT{1.2pt}
              GS-Group (Baseline) & 0.32 & 0.30 \\
            GS-Group+Qwen2.5-VL& 0.78 & 0.77\\
		     +Global Reasoning& 0.89& 0.88\\ 

                \textbf{+Local Refinement}& \textbf{0.95}& \textbf{0.94}\\ 
            \end{tabular}
		\captionsetup{font=footnotesize}
		\subcaption{\textbf{Performance of each component in GLSpaG.}  Adding global reasoning and local refinement progressively improves performance over Qwen2.5-VL.}
		\label{tab:GLGroup}
	\end{subtable}
	\quad
	\renewcommand\arraystretch{1.1}
	\scriptsize
	\begin{subtable}[t]{0.34\textwidth}
		\centering
            \begin{tabular}{ccc}
                Methods& mIoU& mBIoU\\ 
		      \ChangeRT{1.2pt}
		     w/o K-means& 0.38& 0.38\\ 
                K-means+Random&  0.76& 0.75\\ 
                Totally Random& 0.59& 0.58\\ 
               \textbf{ K-means + Top-K-ID (Ours)}& \textbf{0.95}& \textbf{0.94}\\ 
            \end{tabular}
		\captionsetup{font=footnotesize}
		\subcaption{\textbf{Global camera sampling.} K-means view clustering and Top-K ID selection play a crucial role in the global camera sampling process.}
		\label{tab:Global}
	\end{subtable}
	\quad
	\renewcommand\arraystretch{1.1}
	\scriptsize
	\quad
        \begin{subtable}[t]{0.24\textwidth}
		\centering
            \begin{tabular}{cc}
                Method & Speed (FPS)\\ 
		      \ChangeRT{1.2pt}
                \textbf{REALM (Ours)}& \textbf{354.72}  \\ 
                Gaga& 204.49\\ 
                 GS-Group& 305.79\\ 
                GAGS& 107.06\\ 
            \end{tabular}
		\captionsetup{font=footnotesize}
		\subcaption{\textbf{Rendering Efficiency.} The proposed methods do not affect the novel view rendering speed.}
		\label{tab:rendering_speed}
	\end{subtable} \\[1em]
    \\
	\renewcommand\arraystretch{1.2}
	\scriptsize
	\begin{subtable}[t]{0.32\textwidth}
		\centering
            \begin{tabular}{ccc}
                Value of $N^{\text{cluster}}$ & mIoU& mBIoU\\ 
		      \ChangeRT{1.2pt}
               w/o K-means & 0.38& 0.38\\ 
               $N^{\text{cluster}}=2$&  0.76& 0.75\\ 
               \textbf{$N^{\text{cluster}}=24$} & \textbf{0.95}& \textbf{0.94}\\ 
                $N^{\text{cluster}}=128$ & 0.56& 0.56\\ 
            \end{tabular}
		\captionsetup{font=footnotesize}
		\subcaption{\textbf{K-means Clusters.} Both insufficient and excessive clustering can harm multi-view reasoning.}
		\label{tab:k_means_clusters}
	\end{subtable}
	\quad
	\renewcommand\arraystretch{1.2}
	\scriptsize
		\begin{subtable}[t]{0.34\textwidth}
		\centering
            \begin{tabular}{ccc}
                Value of $N^{\text{global}}$& mIoU & mBIoU  \\ 
		      \ChangeRT{1.2pt}
             {  $N^{\text{global}}=4$ }& 0.81& {0.80}\\ 
                $N^{\text{global}}=8$& \textbf{0.95}& \textbf{0.94}\\
                $N^{\text{global}}=16$& 0.95& 0.94\\
                & &
            \end{tabular}
		\captionsetup{font=footnotesize}
		\subcaption{\textbf{Number of global cameras.}  Our method is robust to the hyper-parameter $N^{\text{global}}$.}
	\label{tab:global_camera_number}
        \end{subtable}
	\quad
	\renewcommand\arraystretch{1.2}
	\scriptsize
	\begin{subtable}[t]{0.30\textwidth}
		\centering
            \begin{tabular}{ccc}
                Refinement Steps& mIoU& mBIoU\\ 
		      \ChangeRT{1.2pt}
		    itr=10& 0.94& 0.93\\ 
                \textbf{itr=50}& \textbf{0.95}& \textbf{0.94}\\ 
                itr=500& 0.79& 0.76\\ 
                itr=1000& 0.74& 0.71\\ 
            \end{tabular}
		\captionsetup{font=footnotesize}
		\subcaption{\textbf{Local refinement steps.} Excessive finetuning can lead to overfitting and degradation.}
		\label{tab:refine_itr}
	\end{subtable}
\vspace{-5pt}
\caption{\textbf{Ablation Study.}
We conduct a detailed ablation study on ``\textit{Figurines}'' of the LERF dataset to evaluate the contribution of each component in our method. Cells highlighted in \textbf{bold} indicate the best performance. }
\label{tab:abla}
\end{table*}

\subsection{Main Results}

\noindent\textbf{Qualitative Comparisons.}  Previous methods enable 3D localization by leveraging the language understanding capabilities of CLIP~\cite{radford2021learning} or Grounded-SAM~\cite{ren2024grounded}. While these approaches offer basic open-vocabulary querying capabilities, they lack the ability to perform reasoning over implicit instructions. We visualize the performance between REALM and baselines under different type of implicit queries. The results are presented in Fig.~\ref{fig:vis_lerf}. Demos can be found in the supplementary materials.

\noindent \textit{(1) Spatial Relationship. }
 For example, in the scene `\textit{Teatime}', when given the query `\textit{Which is the drink the teddy bear is holding?}', previous methods tend to focus solely on the keyword `teddy bear' and `drink', resulting in incorrect localization. In contrast, our method finds the drink held by the teddy bear, which is a coffee mug. 
 
 \noindent \textit{(2) Ambiguous Description.} This type of query does not explicitly specify the target object; rather, it describes the object's function or intrinsic attributes. For example, consider the query: ``What is a refreshing beverage made of fruit?'' The model infers that the target object is orange juice.

 \noindent \textit{(2) Contextual Understanding.} This capability requires the model to reason about the target object given a complex context. For example, consider a scenario that everyone else is asleep but you wish to watch videos; REALM observes the scene and selects an earphone as the target object.

\noindent\textbf{Quantitative Comparisons.} We quantitatively evaluate the performance of REALM on both implicit and explicit queries. 
A subset of results is presented in Tab.~\ref{tab:comp}. More results can be found in the supplementary materials.

\noindent \textit{(1) Implicit Queries. }On implicit queries, REALM demonstrates a substantial improvement in performance relative to baseline methods. Previous methods are unable to reason effectively about such queries; even when they correctly identify the target object, they still erroneously activate non-target objects, resulting in performance that is more than 50\% lower on the LERF dataset and more than 35\% lower on the 3D-OVS dataset compared to REALM.

\noindent \textit{(2) Explicit Queries. }The quantitative results for explicit queries are provided in the supplementary materials.



\noindent\textbf{Language-Driven 3D Editing.} With accurate 3D object localization, REALM enables precise and fine-grained scene editing without disturbing surrounding objects. As shown in Fig.~\ref{fig:edit_results}, our model supports a variety of 3D editing tasks, including object removal, replacement, and stylization. REALM modifies the scene without interfering with adjacent content, ensuring faithful preservation of occlusion relationships. In tasks involving large-scale appearance changes, such as stylization, REALM effectively isolates the target object, leaving surrounding regions unaffected.


\subsection{Ablation Study}
We conduct a detailed ablation study of REALM on the ``\textit{Figurines}'' scene from the LERF~\cite{kerr2023lerf} dataset. The results are shown in Tab.~\ref{tab:abla} and Fig.~\ref{fig:random_vs_ours}.

\noindent\textbf{REALM vs. Direct Image Inputs.} Our global stage is crucial for grounding the object. To assess its contribution, we ablate it by simultaneously feeding one or more random views to the MLLM, allowing it to select one single best view, and then running \textit{LMSeg} on that chosen image. We repeat this procedure 10 times and report the statistics. As shown in Fig.~\ref{fig:random_vs_ours}, this strategy is highly sensitive to viewpoint selection, whereas REALM grounds the target object with minimal stochasticity.

\begin{figure}
    \centering
    \includegraphics[width=1.0\linewidth]{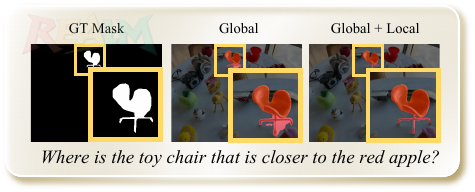}
    
    \caption{\textbf{Ablation study on GLSpaG.} The local grounding stage refines the 3D segmentation results. 
    } 
    \vspace{-10pt}
    \label{fig:glgroup_ab}
\end{figure}

\noindent\textbf{Each component of GLSpaG.}
As shown in Tab.~\ref{tab:GLGroup}, we evaluate the performance after completing each stage of \textit{GLSpaG}. Baseline method (GS-Group) is unable to handle implicit queries. When combined with Qwen2.5-VL, GS-Group can achieve substantial performance improvement. Since result is highly unstable, we report the average score in the table. With our Global Reasoning module, the agent can automatically select appropriate viewpoints for segmentation, leading to a stable mIoU of around 0.89. With the additional {Local Refinement} module, the predicted masks exhibit well-aligned boundaries (see Fig.~\ref{fig:glgroup_ab}).

\noindent\textbf{Global camera sampling strategy.}
The global camera sampling strategy involves two key steps. Firstly, we apply K-means clustering to the training camera poses to ensure diverse viewpoints. Secondly, we select the top-$k$ views that observe the most instances, allowing the model to capture more comprehensive global context. The results in Tab.~\ref{tab:Global} highlight the critical role of each step.

\noindent\textbf{Rendering efficiency.} 
We evaluate the rendering efficiency of REALM, as shown in Tab.~\ref{tab:rendering_speed}. Since our pipeline only renders single-channel masks, it achieves faster rendering speeds compared to other methods.


\noindent\textbf{K-means clusters. }
As shown in Tab.~\ref{tab:k_means_clusters}, the number of clusters ($N^{\text{cluster}}$) significantly affects grounding accuracy. We set $N^{\text{cluster}}{=}24$ in this work, as both too few and too many clusters hinder the selection of optimal global views.

\noindent\textbf{Number of global cameras.} As shown int Tab.~\ref{tab:global_camera_number}, we experiment with different values of $N^{\text{global}}$ and observe that the model remains relatively robust across this range. In practice, we set $N^{\text{global}} = 8$.

\begin{table}
\vspace{-7pt}
    \centering
\resizebox{0.47\textwidth}{!}{
    \renewcommand\arraystretch{1.1}
    \tiny
    \begin{tabular}{lcccc}
    Stage &Calling MLLM (Global) &Calling MLLM (Local) & Local Refine & Total Time\\
    \ChangeRT{0.8pt}
     Time Cost (s)& 2.53 & 2.48 & 3.67 & 8.68
    \end{tabular}
}
\vspace{-10pt}
   \caption{{Inference time analysis.}}
   \label{tab:inference_speed} 
\vspace{-13pt}
\end{table}
\noindent\textbf{Local refinement steps.}
Tab.~\ref{tab:refine_itr} illustrates how the model's performance evolves with an increasing number of local refinement steps. Since refinement is performed on a limited number of views, excessive optimization can lead to overfitting the 3D mask to those specific viewpoints, ultimately degrading the segmentation performance. 

\noindent\textbf{Running Time Analysis.} As shown in Tab.~\ref{tab:inference_speed}, the total time required to process a prompt is less than 10\,s. The operations that invoke the MLLM for each view can be executed in parallel. 
Taking the \textit{``Figurines''} scene as an example, calling the MLLM in the global stage takes 2.53\,s, while the local stage requires 2.48\,s. As for the local refinement stage, the optimization is performed for only 50 iterations, taking 3.67\,s in total.




\section{Conclusion}
We introduced \textbf{REALM}, an MLLM agent framework for open-world 3D reasoning segmentation on 3D Gaussian Splatting. REALM constructs a 3D feature field, performs image-level reasoning with \emph{LMSeg}, and aggregates per-view predictions via the hierarchical \emph{GLSpaG} procedure to obtain robust, fine-grained 3D masks, and it further enables diverse 3D editing operations.
For evaluation, we re-annotate LERF and 3D-OVS with implicit queries and introduce {REALM3D}, a large-scale benchmark covering both reasoning and non-reasoning prompt–mask pairs.
Extensive experiments demonstrate that REALM achieves remarkable performance in 3D segmentation and editing. 



\section*{Acknowledgement}
This work was supported in part by the National Natural Science Foundation of China under Grants (No. 62206082, 62422204, 62502135), the Zhejiang Provincial Natural Science Foundation of China under Grants (No. LRG26F020001, LQN25F030014), the Key Research and Development Program of Zhejiang Province
(No. 2025C01026), the Scientific Research Innovation Capability Support Project for Young Faculty.

{
    \small
    \bibliographystyle{ieeenat_fullname}
    \bibliography{main}

@String(TOG= {ACM Trans. Graph.})

@String(AAAI = {AAAI})

@String(TOG   = {ACM TOG})

@inproceedings{zhu2025vgmamba,
  title={VGMamba: Attribute-to-Location Clue Reasoning for Quantity-Agnostic 3D Visual Grounding},
  author={Zhu, Yihang and Zhang, Jinhao and Wang, Yuxuan and Wu, Aming and Deng, Cheng},
  booktitle={Proceedings of the IEEE/CVF International Conference on Computer Vision},
  pages={5295--5304},
  year={2025}
}

@inproceedings{zhu2024scanreason,
  title={Scanreason: Empowering 3d visual grounding with reasoning capabilities},
  author={Zhu, Chenming and Wang, Tai and Zhang, Wenwei and Chen, Kai and Liu, Xihui},
  booktitle={European Conference on Computer Vision},
  pages={151--168},
  year={2024},
  organization={Springer}
}

@article{feng2025ie,
  title={IE-SRGS: An Internal-External Knowledge Fusion Framework for High-Fidelity 3D Gaussian Splatting Super-Resolution},
  author={Feng, Xiang and Zhong, Tieshi and Chang, Shuo and Wang, Weiliu and Wang, Chengkai and Chen, Yifei and Wang, Yuhe and Kuang, Zhenzhong and Yin, Xuefei and Zhu, Yanming},
  journal={arXiv preprint arXiv:2511.22233},
  year={2025}
}

@article{yang2025sr4d,
  title={SR4D: Dynamic Scene Super Resolution from Monocular Videos},
  author={Yang, Chuxiao and Shi, Changyue and Hu, Xinyuan and Zhu, Suguo and Ding, Jiajun and Wang, Yeru and Tan, Min},
  journal={Knowledge-Based Systems},
  pages={114869},
  year={2025},
  publisher={Elsevier}
}

@article{hu2025srsplat,
  title={SRSplat: Feed-Forward Super-Resolution Gaussian Splatting from Sparse Multi-View Images},
  author={Hu, Xinyuan and Shi, Changyue and Yang, Chuxiao and Chen, Minghao and Ding, Jiajun and Wei, Tao and Wei, Chen and Yu, Zhou and Tan, Min},
  journal={arXiv preprint arXiv:2511.12040},
  year={2025}
}

@article{shi2025sparse4dgs,
  title={Sparse4DGS: 4D Gaussian Splatting for Sparse-Frame Dynamic Scene Reconstruction},
  author={Shi, Changyue and Yang, Chuxiao and Hu, Xinyuan and Chen, Minghao and Pan, Wenwen and Yang, Yan and Ding, Jiajun and Yu, Zhou and Yu, Jun},
  journal={arXiv preprint arXiv:2511.07122},
  year={2025}
}

@article{hu2025texture,
  title={Texture-aware 3D Gaussian Splatting for sparse view reconstructions},
  author={Hu, Xinyuan and Shi, Changyue and Yang, Chuxiao and Chen, Minghao and Gu, Xiaoling and Ding, Jiajun and He, Jifa and Fan, Jianping},
  journal={Applied Soft Computing},
  pages={113530},
  year={2025},
  publisher={Elsevier}
}

@article{fang2022out,
  title={Is out-of-distribution detection learnable?},
  author={Fang, Zhen and Li, Yixuan and Lu, Jie and Dong, Jiahua and Han, Bo and Liu, Feng},
  journal={Advances in Neural Information Processing Systems},
  volume={35},
  pages={37199--37213},
  year={2022}
}

@article{fang2024learnability,
  title={On the Learnability of Out-of-distribution Detection},
  author={Fang, Zhen and Li, Yixuan and Liu, Feng and Han, Bo and Lu, Jie},
  journal={Journal of Machine Learning Research},
  volume={25},
  number={84},
  pages={1--83},
  year={2024}
}

@inproceedings{huang2025reason3d,
  title={Reason3d: Searching and reasoning 3d segmentation via large language model},
  author={Huang, Kuan-Chih and Li, Xiangtai and Qi, Lu and Yan, Shuicheng and Yang, Ming-Hsuan},
  booktitle={International Conference on 3D Vision 2025},
  year={2025}
}

@article{feng2024srgs,
  title={Srgs: Super-resolution 3d gaussian splatting},
  author={Feng, Xiang and He, Yongbo and Wang, Yubo and Yang, Yan and Li, Wen and Chen, Yifei and Kuang, Zhenzhong and Fan, Jianping and Jun, Yu and others},
  journal={arXiv preprint arXiv:2404.10318},
  year={2024}
}

@article{shi2025mmgs,
  title={MMGS: Multi-Model Synergistic Gaussian Splatting for Sparse View Synthesis},
  author={Shi, Changyue and Yang, Chuxiao and Hu, Xinyuan and Yang, Yan and Ding, Jiajun and Tan, Min},
  journal={Image and Vision Computing},
  volume={158},
  pages={105512},
  year={2025},
  publisher={Elsevier}
}

@article{muller2022instant,
  title={Instant neural graphics primitives with a multiresolution hash encoding},
  author={M{\"u}ller, Thomas and Evans, Alex and Schied, Christoph and Keller, Alexander},
  journal={ACM transactions on graphics (TOG)},
  volume={41},
  number={4},
  pages={1--15},
  year={2022},
  publisher={ACM New York, NY, USA}
}

@inproceedings{chen2022tensorf,
  title={Tensorf: Tensorial radiance fields},
  author={Chen, Anpei and Xu, Zexiang and Geiger, Andreas and Yu, Jingyi and Su, Hao},
  booktitle={European conference on computer vision},
  pages={333--350},
  year={2022},
  organization={Springer}
}

@article{tomasi1992shape,
  title={Shape and motion from image streams under orthography: a factorization method},
  author={Tomasi, Carlo and Kanade, Takeo},
  journal={International journal of computer vision},
  volume={9},
  pages={137--154},
  year={1992},
  publisher={Springer}
}

@article{ullman1979interpretation,
  title={The interpretation of structure from motion},
  author={Ullman, Shimon},
  journal={Proceedings of the Royal Society of London. Series B. Biological Sciences},
  volume={203},
  number={1153},
  pages={405--426},
  year={1979},
  publisher={The Royal Society London}
}

@inproceedings{wang2025vggt,
  title={Vggt: Visual geometry grounded transformer},
  author={Wang, Jianyuan and Chen, Minghao and Karaev, Nikita and Vedaldi, Andrea and Rupprecht, Christian and Novotny, David},
  booktitle={Proceedings of the Computer Vision and Pattern Recognition Conference},
  pages={5294--5306},
  year={2025}
}

@article{imp2024,
  title={Imp: Highly Capable Large Multimodal Models for Mobile Devices},
  author={Shao, Zhenwei and Yu, Zhou and Yu, Jun and Ouyang, Xuecheng and Zheng, Lihao and Gai, Zhenbiao and Wang, Mingyang and Ding, Jiajun},
  journal={arXiv preprint arXiv:2405.12107},
  year={2024}
}

@article{fu2023guiding,
  title={Guiding instruction-based image editing via multimodal large language models},
  author={Fu, Tsu-Jui and Hu, Wenze and Du, Xianzhi and Wang, William Yang and Yang, Yinfei and Gan, Zhe},
  journal={arXiv preprint arXiv:2309.17102},
  year={2023}
}

@inproceedings{shi2024language,
  title={Language embedded 3d gaussians for open-vocabulary scene understanding},
  author={Shi, Jin-Chuan and Wang, Miao and Duan, Hao-Bin and Guan, Shao-Hua},
  booktitle={Proceedings of the IEEE/CVF Conference on Computer Vision and Pattern Recognition},
  pages={5333--5343},
  year={2024}
}

@inproceedings{ding2025gpt4image,
  title={GPT4Image: Large Pre-trained Models Help Vision Models Learn Better on Perception Task},
  author={Ding, Ning and Tang, Yehui and Fu, Zhongqian and Xu, Chao and Han, Kai and Wang, Yunhe},
  booktitle={Companion Proceedings of the ACM on Web Conference 2025},
  pages={2056--2065},
  year={2025}
}

@article{raffel2020exploring,
  title={Exploring the limits of transfer learning with a unified text-to-text transformer},
  author={Raffel, Colin and Shazeer, Noam and Roberts, Adam and Lee, Katherine and Narang, Sharan and Matena, Michael and Zhou, Yanqi and Li, Wei and Liu, Peter J},
  journal={Journal of machine learning research},
  volume={21},
  number={140},
  pages={1--67},
  year={2020}
}

@article{brown2020language,
  title={Language models are few-shot learners},
  author={Brown, Tom and Mann, Benjamin and Ryder, Nick and Subbiah, Melanie and Kaplan, Jared D and Dhariwal, Prafulla and Neelakantan, Arvind and Shyam, Pranav and Sastry, Girish and Askell, Amanda and others},
  journal={Advances in neural information processing systems},
  volume={33},
  pages={1877--1901},
  year={2020}
}

@article{liu2023weakly,
  title={Weakly supervised 3d open-vocabulary segmentation},
  author={Liu, Kunhao and Zhan, Fangneng and Zhang, Jiahui and Xu, Muyu and Yu, Yingchen and El Saddik, Abdulmotaleb and Theobalt, Christian and Xing, Eric and Lu, Shijian},
  journal={Advances in Neural Information Processing Systems},
  volume={36},
  pages={53433--53456},
  year={2023}
}

@inproceedings{radford2021learning,
  title={Learning transferable visual models from natural language supervision},
  author={Radford, Alec and Kim, Jong Wook and Hallacy, Chris and Ramesh, Aditya and Goh, Gabriel and Agarwal, Sandhini and Sastry, Girish and Askell, Amanda and Mishkin, Pamela and Clark, Jack and others},
  booktitle={International conference on machine learning},
  pages={8748--8763},
  year={2021},
  organization={PmLR}
}

@inproceedings{liu2025reasongrounder,
  title={ReasonGrounder: LVLM-Guided Hierarchical Feature Splatting for Open-Vocabulary 3D Visual Grounding and Reasoning},
  author={Liu, Zhenyang and Wang, Yikai and Zheng, Sixiao and Pan, Tongying and Liang, Longfei and Fu, Yanwei and Xue, Xiangyang},
  booktitle={Proceedings of the Computer Vision and Pattern Recognition Conference},
  pages={3718--3727},
  year={2025}
}

@inproceedings{lai2024lisa,
  title={Lisa: Reasoning segmentation via large language model},
  author={Lai, Xin and Tian, Zhuotao and Chen, Yukang and Li, Yanwei and Yuan, Yuhui and Liu, Shu and Jia, Jiaya},
  booktitle={Proceedings of the IEEE/CVF Conference on Computer Vision and Pattern Recognition},
  pages={9579--9589},
  year={2024}
}

@inproceedings{li2023blip,
  title={Blip-2: Bootstrapping language-image pre-training with frozen image encoders and large language models},
  author={Li, Junnan and Li, Dongxu and Savarese, Silvio and Hoi, Steven},
  booktitle={International conference on machine learning},
  pages={19730--19742},
  year={2023},
  organization={PMLR}
}

@article{alayrac2022flamingo,
  title={Flamingo: a visual language model for few-shot learning},
  author={Alayrac, Jean-Baptiste and Donahue, Jeff and Luc, Pauline and Miech, Antoine and Barr, Iain and Hasson, Yana and Lenc, Karel and Mensch, Arthur and Millican, Katherine and Reynolds, Malcolm and others},
  journal={Advances in neural information processing systems},
  volume={35},
  pages={23716--23736},
  year={2022}
}

@inproceedings{zhou2024feature,
  title={Feature 3dgs: Supercharging 3d gaussian splatting to enable distilled feature fields},
  author={Zhou, Shijie and Chang, Haoran and Jiang, Sicheng and Fan, Zhiwen and Zhu, Zehao and Xu, Dejia and Chari, Pradyumna and You, Suya and Wang, Zhangyang and Kadambi, Achuta},
  booktitle={Proceedings of the IEEE/CVF Conference on Computer Vision and Pattern Recognition},
  pages={21676--21685},
  year={2024}
}

@inproceedings{kim2024garfield,
  title={Garfield: Group anything with radiance fields},
  author={Kim, Chung Min and Wu, Mingxuan and Kerr, Justin and Goldberg, Ken and Tancik, Matthew and Kanazawa, Angjoo},
  booktitle={Proceedings of the IEEE/CVF Conference on Computer Vision and Pattern Recognition},
  pages={21530--21539},
  year={2024}
}

@article{lyu2024gaga,
  title={Gaga: Group any gaussians via 3d-aware memory bank},
  author={Lyu, Weijie and Li, Xueting and Kundu, Abhijit and Tsai, Yi-Hsuan and Yang, Ming-Hsuan},
  journal={arXiv preprint arXiv:2404.07977},
  year={2024}
}

@article{ren2024grounded,
  title={Grounded sam: Assembling open-world models for diverse visual tasks},
  author={Ren, Tianhe and Liu, Shilong and Zeng, Ailing and Lin, Jing and Li, Kunchang and Cao, He and Chen, Jiayu and Huang, Xinyu and Chen, Yukang and Yan, Feng and others},
  journal={arXiv preprint arXiv:2401.14159},
  year={2024}
}

@article{achiam2023gpt,
  title={Gpt-4 technical report},
  author={Achiam, Josh and Adler, Steven and Agarwal, Sandhini and Ahmad, Lama and Akkaya, Ilge and Aleman, Florencia Leoni and Almeida, Diogo and Altenschmidt, Janko and Altman, Sam and Anadkat, Shyamal and others},
  journal={arXiv preprint arXiv:2303.08774},
  year={2023}
}

@article{liu2023visual,
  title={Visual instruction tuning},
  author={Liu, Haotian and Li, Chunyuan and Wu, Qingyang and Lee, Yong Jae},
  journal={Advances in neural information processing systems},
  volume={36},
  pages={34892--34916},
  year={2023}
}

@article{bai2025qwen2,
  title={Qwen2. 5-vl technical report},
  author={Bai, Shuai and Chen, Keqin and Liu, Xuejing and Wang, Jialin and Ge, Wenbin and Song, Sibo and Dang, Kai and Wang, Peng and Wang, Shijie and Tang, Jun and others},
  journal={arXiv preprint arXiv:2502.13923},
  year={2025}
}

@article{kerr2024robot,
  title={Robot see robot do: Imitating articulated object manipulation with monocular 4d reconstruction},
  author={Kerr, Justin and Kim, Chung Min and Wu, Mingxuan and Yi, Brent and Wang, Qianqian and Goldberg, Ken and Kanazawa, Angjoo},
  journal={arXiv preprint arXiv:2409.18121},
  year={2024}
}

@article{zheng2024gaussiangrasper,
  title={Gaussiangrasper: 3d language gaussian splatting for open-vocabulary robotic grasping},
  author={Zheng, Yuhang and Chen, Xiangyu and Zheng, Yupeng and Gu, Songen and Yang, Runyi and Jin, Bu and Li, Pengfei and Zhong, Chengliang and Wang, Zengmao and Liu, Lina and others},
  journal={IEEE Robotics and Automation Letters},
  year={2024},
  publisher={IEEE}
}

@article{mildenhall2021nerf,
  title={Nerf: Representing scenes as neural radiance fields for view synthesis},
  author={Mildenhall, Ben and Srinivasan, Pratul P and Tancik, Matthew and Barron, Jonathan T and Ramamoorthi, Ravi and Ng, Ren},
  journal={Communications of the ACM},
  volume={65},
  number={1},
  pages={99--106},
  year={2021},
  publisher={ACM New York, NY, USA}
}

@inproceedings{cen2025segment,
  title={Segment any 3d gaussians},
  author={Cen, Jiazhong and Fang, Jiemin and Yang, Chen and Xie, Lingxi and Zhang, Xiaopeng and Shen, Wei and Tian, Qi},
  booktitle={Proceedings of the AAAI Conference on Artificial Intelligence},
  volume={39},
  number={2},
  pages={1971--1979},
  year={2025}
}

@article{peng2024gags,
  title={GAGS: Granularity-Aware Feature Distillation for Language Gaussian Splatting},
  author={Peng, Yuning and Wang, Haiping and Liu, Yuan and Wen, Chenglu and Dong, Zhen and Yang, Bisheng},
  journal={arXiv preprint arXiv:2412.13654},
  year={2024}
}

@inproceedings{qin2024langsplat,
  title={Langsplat: 3d language gaussian splatting},
  author={Qin, Minghan and Li, Wanhua and Zhou, Jiawei and Wang, Haoqian and Pfister, Hanspeter},
  booktitle={Proceedings of the IEEE/CVF Conference on Computer Vision and Pattern Recognition},
  pages={20051--20060},
  year={2024}
}

@inproceedings{kirillov2023segment,
  title={Segment anything},
  author={Kirillov, Alexander and Mintun, Eric and Ravi, Nikhila and Mao, Hanzi and Rolland, Chloe and Gustafson, Laura and Xiao, Tete and Whitehead, Spencer and Berg, Alexander C and Lo, Wan-Yen and others},
  booktitle={Proceedings of the IEEE/CVF international conference on computer vision},
  pages={4015--4026},
  year={2023}
}

@inproceedings{kerr2023lerf,
  title={Lerf: Language embedded radiance fields},
  author={Kerr, Justin and Kim, Chung Min and Goldberg, Ken and Kanazawa, Angjoo and Tancik, Matthew},
  booktitle={Proceedings of the IEEE/CVF International Conference on Computer Vision},
  pages={19729--19739},
  year={2023}
}

@inproceedings{cheng2023tracking,
  title={Tracking anything with decoupled video segmentation},
  author={Cheng, Ho Kei and Oh, Seoung Wug and Price, Brian and Schwing, Alexander and Lee, Joon-Young},
  booktitle={Proceedings of the IEEE/CVF International Conference on Computer Vision},
  pages={1316--1326},
  year={2023}
}

@inproceedings{ye2024gaussian,
  title={Gaussian grouping: Segment and edit anything in 3d scenes},
  author={Ye, Mingqiao and Danelljan, Martin and Yu, Fisher and Ke, Lei},
  booktitle={European Conference on Computer Vision},
  pages={162--179},
  year={2024},
  organization={Springer}
}

@article{kerbl20233d,
  title={3d gaussian splatting for real-time radiance field rendering.},
  author={Kerbl, Bernhard and Kopanas, Georgios and Leimk{\"u}hler, Thomas and Drettakis, George},
  journal={ACM Trans. Graph.},
  volume={42},
  number={4},
  pages={139--1},
  year={2023}
}
}


\end{document}